\documentclass[3p,preprint,times,11pt]{elsarticle}
\usepackage{graphicx}
\usepackage{amsmath}
\usepackage{amssymb}
\usepackage[table,dvipsnames]{xcolor}
\usepackage[most]{tcolorbox}
\usepackage{booktabs}
\usepackage{fontawesome}
\usepackage{hyperref}
\usepackage{pdfpages}

\newcommand{\bu}{\boldsymbol{u}}
\newcommand{\bx}{\boldsymbol{x}}
\newcommand{\bm}{\boldsymbol{m}}
\newcommand{\bw}{\boldsymbol{w}}
\newcommand{\cC}{\mathcal{C}}
\newcommand{\mstrut}{\vphantom{\smin_{j\in{\cal C}_+}}}

\DeclareMathOperator*{\smin}{smin}
\DeclareMathOperator*{\smax}{smax}
\DeclareMathOperator*{\rmin}{rmin}
\DeclareMathOperator*{\rmax}{rmax}

\newtcolorbox{fwbox}{
    enhanced,
    boxrule=0.5pt,
    boxsep=0pt,
    top=-6pt,
    width=.9\linewidth,
    colback = white,
    colframe = Gray!60!white,
    borderline={0.5mm}{0mm}{Gray!60!white},
}

\makeatletter
\def\ps@pprintTitle{%
  \let\@oddhead\@empty
  \let\@evenhead\@empty
  \def\@oddfoot{\reset@font\hfil\thepage\hfil}
  \let\@evenfoot\@oddfoot
}
\makeatother

\begin{document}

\begin{frontmatter}

\title{Fast and Accurate Explanations of Distance-Based Classifiers by Uncovering Latent Explanatory Structures}

\author[bifold,tu]{Florian Bley}
\author[bifold,tu]{Jacob Kauffmann}
\author[eth]{Simon Le\'on Krug}
\author[bifold,tu,korea,mpi]{Klaus-Robert M\"uller}
\author[bifold,charite]{Gr\'egoire Montavon\corref{cor1}}
\ead{gregoire.montavon@charite.de}

\cortext[cor1]{Corresponding author}

\address[bifold]{BIFOLD -- Berlin Institute for the Foundations of Learning and Data, Berlin, Germany}
\address[tu]{Machine Learning Group, Technische Universit\"at Berlin, Germany}
\address[korea]{Department of Artificial Intelligence, Korea University, Seoul, Korea}
\address[eth]{Department of Chemistry and Applied Biosciences, ETH Zurich, Switzerland}
\address[mpi]{Max-Planck Institute for Informatics, Saarbruecken, Germany}
\address[charite]{Charit\'e -- Universit\"atsmedizin Berlin, Germany}

\begin{abstract}
Distance-based classifiers, such as k-nearest neighbors and support vector machines, continue to be a workhorse of machine learning, widely used in science and industry. In practice, to derive insights from these models, it is also important to ensure that their predictions are explainable. While the field of Explainable AI has supplied methods that are in principle applicable to any model, it has also emphasized the usefulness of latent structures (e.g.\ the sequence of layers in a neural network) to produce explanations. In this paper, we contribute by uncovering a hidden neural network structure in distance-based classifiers (consisting of linear detection units combined with nonlinear pooling layers) upon which Explainable AI techniques such as layer-wise relevance propagation (LRP) become applicable. Through quantitative evaluations, we demonstrate the advantage of our novel explanation approach over several baselines. We also show the overall usefulness of explaining distance-based models through two practical use cases.
\end{abstract}


\end{frontmatter}

\section{Introduction}

Distance-based classifiers have long been a workhorse of machine learning (ML). They remain widely used in science and industry, and are often the best available method to learn from tabular data or other well-designed feature representations \cite{Chmiela2017,semnani_OCM}, where deep neural networks occasionally struggle to perform optimally \cite{Borisov2024}. Once a meaningful distance function has been constructed, models such as k-nearest neighbors (KNN) or support vector machines (SVM) can be easily and usually successfully applied due to their small number of hyperparameters and their ability to quickly find the global minimum \cite{Chang2011}, the controlled introduction of nonlinearity into the model, and their regularization properties \cite{DBLP:journals/datamine/Burges98}.

\medskip

Distance-based ML models are also promising tools for data science. Considering the growing trend of combining ML models with Explainable AI (XAI) to infer nonlinear input-output relations in real-world data \cite{DBLP:journals/natmi/BinderBHWHHISHD21, DBLP:journals/mima/ZednikB22}, distance-based models can contribute their unique capabilities to shape the inference task. For example, \cite{brusa23} applied SVM and KNN models together with the SHAP \cite{lundberg_lee_2017} explanation technique to identify time-frequency features that are relevant for industrial bearing fault analysis. Similar approaches combining distance-based models and XAI have been demonstrated for other problems in mechanical engineering \cite{hasan2021}, in chemistry \cite{semnani_OCM}, in finance \cite{khanal2023}, or seismology \cite{PERESAN201887}.

\medskip

So far, most of these works combining distance-based models and XAI have relied on so-called \textit{model-agnostic} XAI approaches such as SHAP, which do not distinguish whether the model is an SVM, a KNN model, a neural network, or else. This generality comes however at a price: While these approaches can be deployed out-of-the-box, their reliance on multiple model evaluations with artificially perturbed input features makes them computationally expensive and potentially inaccurate. In contrast, \textit{model-specific} XAI approaches have demonstrated their advantage in the context of neural networks, with methods such as LRP \cite{Bach2015} leveraging latent representations as part of an orderly and computationally efficient explanation process. However, there has been to the best of our knowledge no attempt to design model-specific XAI techniques that address distance-based classifiers. A challenge with these models is the absence of latent representations to use as a starting point.

\medskip

In this paper, we contribute a novel method for explaining distance-based classifiers. Inspired by the `neuralization-propagation' approach proposed in \cite{DBLP:journals/tnn/KauffmannERMSM24}, our method proceeds in two steps: (1) Rewriting the distance-based classifier as a neural network that exactly reproduces the original model's behavior, but exposes latent structures useful for explanation. (2) Designing an LRP attribution procedure that leverages these latent structures and explains the model's prediction in terms of its input features. Our novel method is instantiated to two types of distance-based classifiers, KNNs and SVMs, and can be extended further to regression models (cf.\ Section \ref{sec:dpm}). Compared to model-agnostic methods based on many function evaluations, our method is significantly faster, requiring the equivalent of only two such evaluations. Furthermore, our method does not require any retraining and explains the original unmodified model.

\medskip

Through quantitative evaluations, we show that our method outperforms common model-agnostic XAI methods in terms of explanation accuracy across a broad range of datasets and ML models. Lastly, we demonstrate on two use cases (a wine tasting and a quantum chemistry dataset) the utility of our proposed explainable distance-based models in unveiling complex nonlinear relationships in the data, beyond the reach of linear models or correlation analyses.

\section{Related Work}

When considering the field of Explainable AI, specifically the extraction of nonlinear relationships in the data through an ML model, we can distinguish between three general approaches. We briefly discuss them below together with references to corresponding works in the literature.

\subsection{Intrinsically Interpretable ML Models}
\label{section:relatedwork-intrinsic}

The first approach involves developing an ML architecture that is inherently explainable. Linear models \cite{hastie1990generalized} or generalized additive models (GAMs) \cite{caruana2015} fulfill that description by summing over input features or nonlinear transformations of these features. In these models, the contribution of each input feature can be identified directly from the summation terms. Similar ideas can be found in image classification, where summing operations known as `global average pooling' are placed at the output of convolutional layers \cite{Zhou_2016_CVPR, DBLP:conf/iclr/BrendelB19}. The largest contributors in the pool can then be interpreted as the most important spatial locations in the image. While GAMs and other intrinsically interpretable ML models facilitate the formulation of the explanation problem, their constrained structure limits the scope of nonlinear relationships that can be extracted from the data, potentially leading to limited insights.

\subsection{Model-Agnostic XAI Techniques}
\label{section:relatedwork-agnostic}

Another approach called `model-agnostic' seeks to extract a broader set of nonlinear relations by applying to any nonlinear ML model. Approaches such as Integrated Gradients (IG) \cite{sundararajan2017} view the ML model as a function and explain by repeatedly evaluating the function and its gradient along some path in the input domain. Feature contributions are then derived from the collected gradients. Occlusion-based methods \cite{Zeiler2014} explain by testing the effect of removing each input feature from the input data on the model output. A popular occlusion-based approach is the SHAP method \cite{lundberg_lee_2017} which is theoretically grounded in the framework of Shapley values \cite{shapley1953} and accounts for the effect of removing multiple features simultaneously. A commonality of most model-agnostic XAI techniques is that they all measure effects through local evaluations of the model and its gradient. Consequently, they face a trade-off, where performing more evaluations improves explanation accuracy but increases computational cost (cf.\ \cite{samek2021explaining, FELDMANN2022}).

\subsection{Model-Specific XAI Techniques}
\label{section:relatedwork-specific}

A further approach, aiming to overcome the limitations of model-agnostic XAI techniques, is to leverage specific internal structures present in popular ML models, for example, the hidden layers of a neural network. One example of this approach is the Layer-wise Relevance Propagation (LRP) technique \cite{Bach2015}. LRP generates explanations with a single backward pass through the model, informed by the activations and nonlinearities found at each layer. The LRP approach has been extended to a broad range of neural network architectures, including Transformers \cite{DBLP:conf/icml/AliSEMMW22} and Mamba \cite{DBLP:conf/nips/JafariMME24}. In \cite{DBLP:journals/pr/KauffmannMM20} a latent distance-pooling structure was uncovered in distance-based anomaly models, allowing the LRP approach to be mapped to these non-neural network models. The same principle was further extended in \cite{DBLP:journals/tnn/KauffmannERMSM24} where similar latent structures were uncovered in k-means clustering models. Among non-LRP approaches, we find efficient Shapley value algorithms derived for specific models such as SVMs operating on binary input features \cite{Mastropietro2023} or decision trees \cite{DBLP:journals/natmi/LundbergECDPNKH20}. While these model-specific XAI approaches enable fast and accurate explanations for their respective model classes, they are not applicable to the distance-based classifiers that are the focus of this work.

\section{Uncovering Explanatory Structures in Distance-Based Classifiers}
\label{section:neuralization}

Explainable AI has been particularly successful in explaining the predictions of complex, state-of-the-art neural networks. This success can be attributed to the presence of intermediate representations in these models, serving as explanation structures that can be leveraged in the process of identifying relevant features \cite{Bach2015, DBLP:journals/ijcv/SelvarajuCDVPB20, DBLP:conf/icml/AliSEMMW22, DBLP:journals/natmi/AchtibatDEBWSL23, DBLP:conf/nips/JafariMME24}. However, these structures appear to be absent in distance-based classifiers, which are the focus of this work.
We contribute by showing that distance-based classifiers can be reformulated as equivalent neural networks. These neural networks we uncover are unique in that they provide the necessary structures to facilitate explanation.

\subsection{A Neural Network Equivalent of SVMs}

\begin{figure*}[t!]
    \centering
    \makebox[\textwidth][c]{
    \includegraphics[width=.95\textwidth]{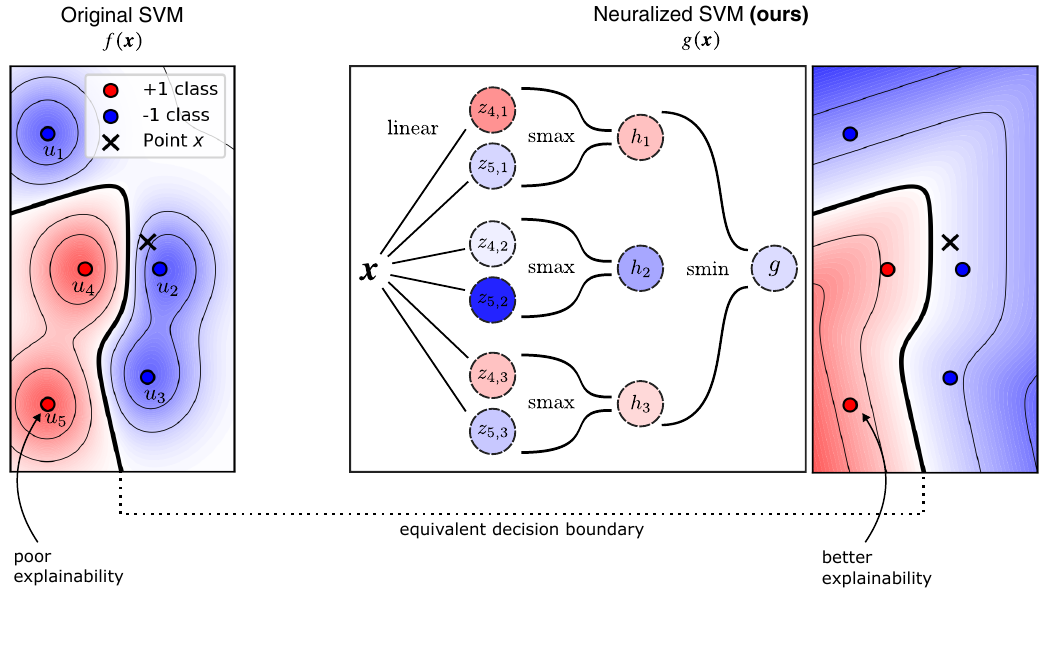}
    }\vspace{-8mm}
    \caption{Proposed neural network reformulation of the SVM to enhance its explainability. \textit{Left:} output $f(\bx)$ of a Gaussian SVM trained on two-dimensional data. Red indicates positive and blue indicates negative. The function $f(\bx)$ saturates near the training data. \textit{Middle:} diagram of the neuralized SVM comprising one linear layer followed by two pooling layers. The neurons are color-coded by activation for a chosen test point (`$\times$' in left panel). \textit{Right:} output $g(\bx)$ of the neuralized model. It maintains the original SVM's decision boundary while eliminating saturation and revealing a piecewise linear structure useful for explanation.}
\label{fig:overview}
\end{figure*}

SVM models  \cite{Cortes1995, muller2001introduction,scholkopf1999advances} classify data points by weighting their similarity to instances of the two classes through a kernel function, here assumed to be Gaussian. The SVM selects a subset of relevant training data -- the support vectors (SVs) -- and weights them using the dual-parameters $\alpha$ found during model optimization. A trained kernel SVM with a Gaussian kernel classifies a data point $\bx$ according to the sign of the function:
\begin{align}
f(\bx) = \sum_{\ell=1}^n y_\ell\cdot\alpha_\ell \exp(-\gamma \cdot \lVert \bx - \bu_\ell \rVert^2) + \theta
\label{eq:svm-boundary}
\end{align}
where $\gamma > 0$ is the kernel smoothness parameter,  $\theta \in \mathbb{R}$ is the model bias and $\bu_\ell$ are the SVs with their class labels $y_\ell \in \{-1, +1\}$. An example of the function $f(\bx)$ is plotted in Figure \ref{fig:overview} (left). The plot highlights local extrema formed near the training data, with the gradient often not pointing towards the decision boundary and thereby failing to support explainability.

To overcome this limitation, we propose an equivalent neural network that incorporates explainable hierarchical structures. Our proposed network is composed of three layers and preserves exactly the original decision boundary provided by Eq.\ \eqref{eq:svm-boundary}:
\begin{center}
\begin{fwbox}
\begin{subequations}\label{eq:svm_neuralization}
\begin{align}
z_{ij} &= (\bx - \bm_{ij})^\top \bw_{ij} + b_{ij} \mstrut
\label{eq:svm_neuralization-1}\\
h_{j} &= {\smax_{i \in \cC_+}}^{\gamma} \big\{z_{ij}\big\}
\label{eq:svm_neuralization-2}\\
g(\bx) &= {\smin_{j \in \cC_-}}^{\gamma} \big\{{h_j }\big\}
\label{eq:svm_neuralization-3}
\end{align}
\end{subequations}
\end{fwbox}
\end{center}
The neural network is depicted in Figure \ref{fig:overview} (middle). The first layer, interpretable as a detection layer, comprises neurons indexed by pairs of opposite-class SVs, where each neuron computes a linear activation with a weight vector $\bw_{ij} = 2 (\bu_i - \bu_j)$, a bias $b_{ij} = \gamma^{-1} \log (\alpha_i/\alpha_j)$ and the SV midpoint $\bm_{ij} = (\bu_i + \bu_j)/2$. These activations are then passed into two more layers, which can be interpreted as pooling layers: a smooth maximum ($\smax^{\gamma} \{a\}= \gamma^{-1} \log\sum\exp\{\gamma a\}$) over positive-class SVs, followed by a smooth minimum ($\smin^{\gamma} \{a\}= \smax^{-\gamma}\{a\}$) over negative-class SVs. To incorporate the model bias $\theta$ we need to add the term $z_{0} = -\gamma^{-1}\log(\theta)$ to the $\cC_+$ pool if the bias is positive, and to the $\cC_-$ pool if the bias is negative. Crucially, in addition to the uncovered linear and pooling explanation structures, one can show that $f(\bx)$ and $g(\bx)$ always have the same sign and thus implement the same decision boundary (cf.\ Supplementary Note A).
Our neural network reformulation can be distinguished from other SVM reformulations (e.g.\ RBF networks \cite{DBLP:journals/neco/MoodyD89} or random feature models \cite{rahimi2007random}) by its focus on supporting explainability. The linear structures of our neuralized SVM are easily identifiable from the neuralized SVM's parameters, and their alignment to the decision boundary can be clearly seen in the plotting of $g(\bx)$ in Figure \ref{fig:overview} (right).

\subsection{A Neural Network Equivalent of KNNs}

We now extend our neuralization strategy to KNN classifiers. These classifiers collect the subset of the $k$ training points closest to the test data point and classify it according to its neighbor's majority class, that is, according to the sign of the function:
\begin{align}
    f(\bx) = \sum_{\ell=1}^n y_\ell \cdot\mathbf{1}_{\{\|\bx-\bu_\ell\|\,\leq\, d_k(\bx)\}}\label{eq:knn-boundary}
\end{align}
where $\bu_\ell$ denote training input samples with their class labels $y_\ell\in\lbrace -1,+1\rbrace$ and $d_k(\bx) = \|\bx-\bu_{k}\|$ is the distance to the $k$-th nearest neighbor. The hyperparameter is assumed to be an odd number, i.e.\ $k=2q-1$ with $q\in\mathbb{Z}^+$. We note that Eq.\ \eqref{eq:knn-boundary} is especially difficult to explain using classical methods due to its non-differentiability; in particular, gradient-based explanations are inapplicable. We propose to restate the KNN decision boundary as the following three-layer network:
\begin{center}
\begin{fwbox}
\begin{subequations}
\begin{align}
z_{ij} &= (\bx-\bm_{ij})^\top \bw_{ij}\mstrut\\
h_j &= {\rmax_{i \in \mathcal{C}_+}}^{q} \{ z_{ij} \}\\
g(\bx) & =  {\rmin_{j \in \mathcal{C}_-}}^{q} \big\{h_j \big\} \label{eq:knn_neuralization}
\end{align}
\end{subequations}
\end{fwbox}
\end{center}
As in the SVM neuralization, the first layer consists of linear detection units indexed by pairs of opposite-class neighbors. Each unit computes a pre-activation $z_{ij}$ where $\bm_{ij}$ and $\bw_{ij}$ are defined analogously to Eq.\ \ref{eq:svm_neuralization} and contains directional information contrasting points of opposite classes. The resulting activations are again passed through pooling layers, this time, a ranked maximum over positive-class neighbors followed by a ranked minimum over negative-class ones, and $\rmin$/$\rmax$ are ranked minimum/maximum operators, i.e.\ $\rmin_{i\in{\cal I}}^q\lbrace a_i\rbrace$ returns the $q$-smallest value from the set $\lbrace a_i\rbrace_{i\in{\cal I}}$. The proposed neural network implements the same decision boundary as the original KNN model (cf.\ Supplementary Note B for a proof), and brings additional linear/pooling explanatory structures, which we will leverage in Section \ref{section:explanation} to produce feature-wise explanations.

\section{LRP Explanations for SVMs and KNNs}
\label{section:explanation}

In this section, we propose a procedure based on layer-wise relevance propagation (LRP) \cite{Bach2015} to accurately and efficiently attribute the predictions of SVMs and KNNs to their input features. Our method takes as a starting point the neural network equivalent of these models uncovered in Section \ref{section:neuralization}, and proceeds by reverse-propagating the output of these models through their multiple layers, until the input features are reached.

\subsection{Propagation in the Pooling Layers}
\label{section:pooling}

To propagate through the neuralized SVM's pooling layers, we follow the approach of \cite{DBLP:journals/pr/KauffmannMM20, DBLP:journals/tnn/KauffmannERMSM24}, where the output of the model is redistributed to the pooling inputs $z_{ij}$ through the LRP rule:
\begin{align}
\label{eq:SVM_pooling}
    R_{ij} &= {\overbrace{\frac{\exp(\beta z_{ij})}{\sum_{i'} \exp(\beta z_{i'j})}}^{\color{red!50!black}p_i}} \cdot {\overbrace{\frac{\exp(-\beta h_j)}{\sum_{j'} \exp(-\beta h_{j'})}}^{\color{blue!50!black}p_j}} \cdot g(\bx)
\end{align}
The terms $p_i$ and $p_j$ are `softargmax' and `softargmin' functions, mirroring the softmax and softmin functions in the forward pass. The hyperparameter $\beta \in \mathbb{R}^+$ controls the `stiffness' of the redistribution. For $\beta$ large, the redistribution becomes an argmax and argmin, and for $\beta$ small, it becomes more uniform. Interestingly, the first term in Eq.\ \eqref{eq:SVM_pooling} can be shown to be equivalent for all $j$, and thus can be denoted by the factor $p_i$ without index $j$ (cf.\ Supplementary Note C).

For the neuralized KNN's pooling layers, we first recall that classification is determined solely by the two $q=(k+1)\slash2$-th closest neighbors in both classes. Similar to the SVM case, we propose a smooth relevance distribution not only to the $q$-th closest data point but also to a neighborhood of indices surrounding it. Specifically, we introduce a parameter \(\kappa\) to define a ``band of indices'' centered on the $q$-th neighbor, which we denote as $\mathcal{P}$ and $\mathcal{N}$ for the positive and negative data points (cf.\ Supplementary Note C for a formal specification). We then propose to redistribute uniformly across these indices according to the LRP rule:
\begin{align}\label{eq:KNN_pooling}
    R_{ij} &=  {\overbrace{\frac{\mathbf{1}_{\{i\,\in\,\mathcal{P}\}}}{\sum_{i'} \mathbf{1}_{\{i'\,\in\,\mathcal{P}\}}}}^{\color{red!50!black}p_i}} \cdot
    {\overbrace{\frac{\mathbf{1}_{\{j\,\in\,\mathcal{N}\}}}{\sum_{j'} \mathbf{1}_{\{j'\,\in\,\mathcal{N}\}}}}^{\color{blue!50!black}p_j}} \cdot g(\bx)
\end{align}
which shares the same structure as Eq.\ \eqref{eq:SVM_pooling} describing the SVM case, namely, the output of the neuralized model multiplied by probability scores $p_i$ and $p_j$.

\subsection{Propagation in the Detection Layer}
\label{section:linear}

As a preliminary step, we analyze the structure of the relevance $R_{ij} = p_i p_j \cdot g(\bx)$ extracted in Eqs.\ \eqref{eq:SVM_pooling} and \eqref{eq:KNN_pooling}. All three terms depend on $\bx$ but in complex nonlinear ways. This makes it difficult to propagate this quantity to the input features in an appropriate manner. To address this difficulty, we adopt a strategy similar to \cite{DBLP:journals/tnn/KauffmannERMSM24, montavon2017dtd}, consisting of building a `relevance model'. Specifically, we use the following local approximation of $R_{ij}$:
\begin{align}
\widehat{R}_{ij} = [p_i p_j]_\text{cst} \cdot z_{ij}
\label{eq:surrogate}
\end{align}
where $z_{ij}$ linearly depends on $\bx$ and $p_i$ and $p_j$ are treated as locally constant (cf.\ Supplementary Note D for a detailed analysis of this approximation, including when the SVM offset $\theta$ is non-zero). The approximation is especially accurate when one pair of support vectors $(i,j)$ strongly dominates the pool, i.e.\ $p_i p_j \approx 1$, or when same-class neighbors are close and the distance between classes is large.

Having defined the relevance model $\widehat{R}_{ij}$ in Eq.\ \eqref{eq:surrogate}, we now proceed with redistributing the classifier's evidence to the input features. For this, we consider the first-order Taylor expansion $\widehat{R}_{ij}(\bx) = \widehat{R}_{ij}(\widetilde{\bx}) + (\bx - \widetilde{\bx})^\top (\bw_{ij} \cdot p_i p_j)$, where $\widetilde{\bx}$ is a reference point and where elements of the dot product identify the amount to be redistributed to each input feature. We propose to choose $\widetilde{\bx}$ on the segment $[\boldsymbol{0},\bm_{ij}]$ which allows us to express the overall redistribution to the input features in closed form as:
\begin{align}
\label{eq:LRP}
\mathcal{E}_\eta(\bx)
&= \sum_{i \in \cC_+}\sum_{j \in \cC_-} (\bx - \eta \cdot \bm_{ij}) \odot (\bw_{ij} \cdot p_i p_j)
\end{align}
where $\eta \in [0,1]$ is a hyperparameter that controls the location of the reference points on their respective segments. Choosing $\eta$ large (setting $\widetilde{\bx}$ close to $\boldsymbol{m}_{ij}$) leads to explanations that are well-contextualized and adjust well to sharp discontinuities in the model's response. Choosing $\eta$ small (setting $\widetilde{\bx}$ close to $\boldsymbol{0}$) is more suited for smoother decision boundaries resulting from subtle interactions between multiple competing detection units. In practice, our experiments will demonstrate that intermediate values of $\eta$ produce the most accurate explanations.

\subsection{Computational Cost}

A direct computation of Eq.\ \eqref{eq:LRP} would require summing over all pairs of data points, and thus introduce a computational complexity quadratic in the number of neighbors or support vectors. However, it is useful to observe that the explanation for any choice of $\eta$ can be expressed as a linear combination of the two edge cases, namely, $\mathcal{E}_\eta = (1-\eta) \cdot \mathcal{E}_{\eta=0} + \eta \cdot \mathcal{E}_{\eta=1}$, and that for these two edge cases, the nested sums can be broken into two separate sums (cf.\ Supplementary Note E). This implies that the explanation cost of our method is \textit{linear} in the number of support vectors or nearest neighbors. Specifically, the computational cost becomes equivalent to only two evaluations of the original classifier. This makes our method much more efficient than other explanation techniques based on repeated function evaluations, such as Integrated Gradients, Occlusion, or SHAP.

\section{Evaluating Explanation Accuracy}

We have proposed in Sections \ref{section:neuralization} and \ref{section:explanation} an approach combining neuralization and propagation via LRP rules to explain two families of distance-based classifiers: kernel SVMs and KNNs. To evaluate our approach (referred in the following as `LRP-SVM' and `LRP-KNN'), we compare explanation accuracy against a number of baselines and across diverse classification tasks and models. A subsequent ablation study examines how the two components of our approach (neuralization and propagation) each contribute to improve explanation accuracy. Qualitative examples of explanations our method produces are provided in Sections \ref{sec:wine_quality} and \ref{sec:dpm}.

\subsection{Evaluation Metric}

Accuracy of an explanation is commonly understood as its ability to pinpoint features that truly contribute to the model's output. To quantify this, we rely on a technique known as ``pixel-flipping'' \cite{Bach2015, samek2017evaluating} which removes input features in decreasing order of importance according to the explanation and monitors the extent to which these removals impact the classifier's output\footnote{Following the pixel-flipping definition in \cite{DBLP:journals/tnn/KauffmannERMSM24}, the set of features removed are inpainted using a kernel density estimator (KDE), which resamples their values conditioned on the remaining features.}. Specifically, after each feature removal step, we record the classification outcome ($+1$: original classification, $-1$: flipped classification), a process that can be represented as a flipping-curve (FC). The explanation accuracy is quantified as the area under this flipping curve (AUFC), and averaged over all data points. The lower the AUFC, the better the explanation. In our setting, AUFC values range from $-1$ to $1$.

\subsection{Datasets and Models}

We perform our evaluation on a diverse set of classification datasets across multiple domains. We provide dataset URLs in Supplementary Note F. We use four binary tabular classification datasets from the UCI Machine Learning Repository: Breast Cancer, Diabetes Risk, Raisin, and Rice. We also include two binary image classification tasks from the MedMNIST benchmark: BreastMNIST and PneumoniaMNIST. Additionally, we adapt the UCI Car Evaluation dataset into a binary task by predicting whether a car is rated as ``unacceptable.'' Furthermore, we derive binary classification tasks from four regression datasets. For the UCI Wine Quality dataset, we classify wines as high or low quality based on the median score. In the UCI Concrete Compressive Strength dataset, we classify samples by whether their compressive strength exceeds the 90th percentile. For the UCI Real Estate Valuation dataset, we distinguish between properties with values above or below the median. Lastly, for the OCM catalyst dataset \cite{dataset_catalyst}, we follow the classification task proposed in \cite{semnani_OCM}, which separates samples based on whether the $\text{C}_2$ yield rate exceeds 13\% and we replicate their feature selection procedure.

A uniform preprocessing is applied to all datasets. The data are shuffled and standardized to zero mean and unit variance. A 20\% validation split is set aside, with 300 examples further reserved for evaluating explanations. The remaining data are used for training. Afterward, the data are rescaled so that the median pairwise distance of training data is one, ensuring comparable $\gamma$ values across datasets and aligning $\gamma=1$ with the median heuristic \cite{sriperumbudur2009}.

We train a Gaussian SVM and a KNN model on the training split for each dataset, optimizing $\gamma$ and $C$ for the SVM and $k$ for KNN via cross-validation. Explanations are generated for all examples in the explanation evaluation set.

\subsubsection{Hyperparameter Selection and Baseline Methods}

We recall from Section \ref{section:explanation} that our method is based on LRP rules, which possess hyperparameters: $\eta$ controls the level of contextualization of the explanation in the detection layer, whereas $\kappa$ and $\beta$ determine how localized the explanation is in terms of the nearest data points or support vectors. Based on empirical analysis, we find that these parameters can be suitably set based on the model's scale parameters $\gamma$ and $k$ indicating the level of nonlinearity. Our empirical analysis is provided in Supplementary Note G, from which we derive the heuristics shown in Table \ref{tab:heuristic}.

\begin{table}[t!]

\newcolumntype{C}[1]{>{\centering\let\newline\\\arraybackslash\hspace{0pt}}m{#1}}

\caption{Proposed heuristics for setting LRP hyperparameters ($\eta$, $\beta$ and $\kappa$) at each layer and for each model, based on model characteristics ($\gamma$ and $k$). The heuristics for SVMs assume that the input data are normalized such that the median Euclidean distance between samples is one.  These heuristics are used in our evaluations in Section \ref{section:results}.}
\label{tab:heuristic}
\centering \medskip \small
\begin{tabular}{C{1.5cm}C{5cm}C{5cm}}
\toprule
 & \multicolumn{2}{c}{LRP hyperparameters}\\
 & in linear layer & in pooling layer \\
\midrule
SVM &
$\eta = \text{median} \big\{0, 0.4\,\log_{10} \gamma + 0.4,1\big\}$ &
$\beta = \gamma$ \\
KNN &
$\eta = 0.8$ &
$\kappa = (k - 1)/2$ \\
\bottomrule
\end{tabular}
\end{table}

Our method is tested against several established explanation baselines. Firstly, we consider Gradient$\,\times\,$Input \cite{shrikumar2017learning} evaluated using the input gradient of the original model. Furthermore, we consider Integrated Gradients \cite{sundararajan2017} using the data origin as a reference point and performing 10 integration steps. Among the gradient-based explanation baselines, we also consider Sensitivity Analysis \cite{zurada1995, baehrens2010explain}, which explains the SVM model output with the squared input gradient. As non-gradient-based baselines, we consider an Occlusion analysis \cite{Zeiler2014}, setting independent features to the data mean and Shapley value sampling \cite{lundberg_lee_2017} (SHAP) using 10 random feature removal sequences. 

\subsection{Results}
\label{section:results}

Results of our evaluation procedure are presented in Table \ref{table:pf}.
\begin{table*}[t!]
\caption{AUFC results for different explanation methods for SVM and KNN. Lower values are better. The best explanation per dataset is highlighted in bold. The second-best explanation is underlined. $^\text{\color{gray}\faHourglassStart}$\ indicates methods with higher computational cost.}
\medskip
\centering \footnotesize
\makebox[\textwidth][c]{
\begin{tabular}{llccccccc}
\toprule
Dataset & Model & GI & IG$^\text{\color{gray}\faHourglassStart}$ & $(\nabla f)^2$ & Occlusion$^\text{\color{gray}\faHourglassStart}$ & SHAP$^\text{\color{gray}\faHourglassStart}$ & LRP-SVM & max std.  \\[-1mm]
& & & & & & & \textbf{(ours)}\\
\midrule
BreastMNIST & SVM $\gamma=3.0$ & $0.222$ & $0.338$ & $0.704$ & $\underline{0.185}$ & $0.345$ & $\mathbf{0.040}$ & $\pm \mathit{0.065}$ \\
Breast Cancer & SVM $\gamma=0.01$ & $0.505$ & $\underline{0.505}$ & $0.703$ & $0.505$ & $0.505$ & $\mathbf{0.503}$ & $\pm \mathit{0.036}$ \\
Car Evaluation & SVM $\gamma=3.0$ & $0.203$ & $\underline{0.035}$ & $0.305$ & $0.090$ & $\mathbf{0.023}$ & $0.058$ & $\pm \mathit{0.021}$ \\
Catalyst & SVM $\gamma=3.0$ & $-0.043$ & $-0.047$ & $0.606$ & $\mathbf{-0.186}$ & $-0.040$ & $\underline{-0.116}$ & $\pm \mathit{0.077}$ \\
Concrete & SVM $\gamma=3.0$ & $0.718$ & $0.678$ & $0.738$ & $0.691$ & $\underline{0.678}$ & $\mathbf{0.655}$ & $\pm \mathit{0.031}$ \\
Diabetes Risk & SVM $\gamma=3.0$ & $0.351$ & $0.352$ & $0.634$ & $\underline{0.318}$ & $0.336$ & $\mathbf{0.234}$ & $\pm \mathit{0.048}$ \\
PneumoniaMNIST & SVM $\gamma=3.0$ & $0.589$ & $0.586$ & $0.884$ & $\underline{0.494}$ & $0.587$ & $\mathbf{0.354}$ & $\pm \mathit{0.039}$ \\
Raisin & SVM $\gamma=0.3$ & $0.646$ & $\underline{0.606}$ & $0.749$ & $0.617$ & $\mathbf{0.601}$ & $0.611$ & $\pm \mathit{0.020}$ \\
Real Estate Valuation & SVM $\gamma=10.0$ & $0.573$ & $0.428$ & $0.543$ & $0.388$ & $\mathbf{0.383}$ & $\underline{0.385}$ & $\pm \mathit{0.044}$ \\
Rice & SVM $\gamma=1.0$ & $0.706$ & $0.715$ & $\mathbf{0.647}$ & $\underline{0.686}$ & $0.709$ & $0.696$ & $\pm \mathit{0.016}$ \\
Wine Quality & SVM $\gamma=10.0$ & $0.460$ & $0.355$ & $0.559$ & $0.359$ & $\underline{0.326}$ & $\mathbf{0.277}$ & $\pm \mathit{0.020}$ \\
\bottomrule
\\[-1mm]
\toprule
Dataset & Model & GI & IG$^\text{\color{gray}\faHourglassStart}$ & $(\nabla f)^2$ & Occlusion$^\text{\color{gray}\faHourglassStart}$ & SHAP$^\text{\color{gray}\faHourglassStart}$ & LRP-KNN  & max std.\\[-1mm]
& & & & & & & \textbf{(ours)}\\
\midrule
BreastMNIST & KNN $k=9$ & - & - & - & $\underline{0.718}$ & $0.728$ & $\mathbf{0.267}$ & $\pm \mathit{0.061}$ \\
Breast Cancer & KNN $k=9$ & - & - & - & $0.804$ & $\underline{0.618}$ & $\mathbf{0.458}$ & $\pm \mathit{0.039}$ \\
Car Evaluation & KNN $k=5$ & - & - & - & $0.222$ & $\mathbf{0.067}$ & $\underline{0.111}$ & $\pm \mathit{0.022}$ \\
Catalyst & KNN $k=5$ & - & - & - & $0.559$ & $\underline{0.487}$ & $\mathbf{0.369}$ & $\pm \mathit{0.072}$ \\
Concrete & KNN $k=9$ & - & - & - & $0.792$ & $\underline{0.791}$ & $\mathbf{0.751}$ & $\pm \mathit{0.027}$ \\
Diabetes Risk & KNN $k=3$ & - & - & - & $0.517$ & $\underline{0.413}$ & $\mathbf{0.278}$ & $\pm \mathit{0.047}$ \\
PneumoniaMNIST & KNN $k=7$ & - & - & - & $0.880$ & $\underline{0.850}$ & $\mathbf{0.189}$ & $\pm \mathit{0.040}$ \\
Raisin & KNN $k=19$ & - & - & - & $0.705$ & $\underline{0.632}$ & $\mathbf{0.623}$ & $\pm \mathit{0.022}$ \\
Real Estate Valuation & KNN $k=3$ & - & - & - & $0.437$ & $\underline{0.425}$ & $\mathbf{0.396}$ & $\pm \mathit{0.040}$ \\
Rice & KNN $k=25$ & - & - & - & $0.727$ & $\mathbf{0.661}$ & $\underline{0.683}$ & $\pm \mathit{0.014}$ \\
Wine Quality & KNN $k=25$ & - & - & - & $0.457$ & $\underline{0.426}$ & $\mathbf{0.393}$ & $\pm \mathit{0.023}$ \\
\bottomrule
\end{tabular}
}
\label{table:pf}
\end{table*}
Our method demonstrates superior explanation accuracy across model types and datasets. For SVMs, our LRP-SVM method ranks best or second-best in 8 out of 11 datasets, with particularly strong improvements in AUFC on high-$\gamma$ (highly nonlinear) models where the local extrema problem in input space is most pronounced. In these cases, the AUFC difference to the second-best method is typically a multiple of the estimated standard deviation. For KNN classifiers, where gradient-based methods are inapplicable, our LRP-KNN method outperforms the Occlusion and SHAP baselines in 9 out of 11 datasets with highly significant AUFC differences. While Occlusion and SHAP perform considerably worse on KNN than SVMs -- likely because single Occlusion steps cannot modify the discrete KNN output -- LRP maintains strong performance through the directedness introduced by our neuralization's detection units. LRP excels on high-dimensional datasets like MNIST variants, where single-feature occlusions are limited to local probing. Overall, SHAP ranks second due to its ability to take larger steps to probe the input space and potentially circumvent local extrema. While SHAP and Occlusion occasionally outperform LRP for SVMs, these explainability gains are small and come at substantially higher computational cost.

\begin{figure*}[t!]
    \centering
    \makebox[\textwidth][c]{\includegraphics[width=1.05\textwidth]{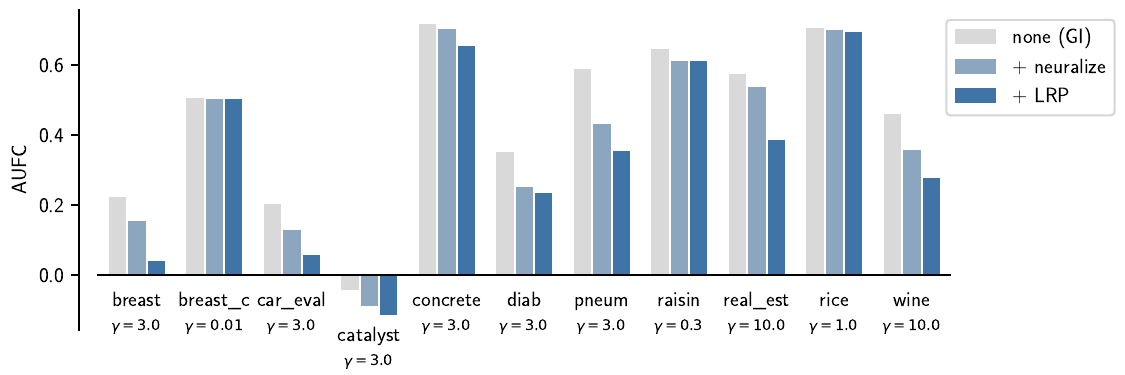}}
    \caption{Ablation study in which the neuralization and propagation steps of our LRP-SVM method are introduced one after another. The results are shown for the same datasets and SVM models as in Table \ref{table:pf}. We observe that both neuralization and propagation contribute to improve explanations, as measured by a reduction in AUFC score.
    }
   \label{fig:ablation}
\end{figure*}

\paragraph{Ablation study} To analyze the individual impact of the two components of our method, i.e.\ (1) neuralization and (2) propagation via LRP rules, we conduct an ablation study in which these two components are introduced successively for each pair of datasets and SVM models. Our starting point is basic Gradient$\,\times\,$Input (GI) applied to the original function $f(\boldsymbol{x})$. The same GI is then performed on the neuralized model $g(\boldsymbol{x})$. Finally, GI is replaced by our propagation approach with the proposed LRP rules. The results are shown in Figure \ref{fig:ablation}.
We observe that introducing neuralization alone improves AUFC scores consistently across all datasets/SVM pairs.  This demonstrates that the desaturation that neuralization incurs on the decision function, by removing spurious local extrema near the training data, is effective in improving explanation quality on its own. Adding the propagation step (i.e.\ our purposely defined LRP rules) further improves explanation quality as measured by a further reduction in AUFC scores. Similar to the results from Table \ref{table:pf} we see a relation between the SVM's parameter $\gamma$ and the extent of AUFC decrease, with highest improvements obtained for large $\gamma$ values. Additionally, we can observe that both neuralization and propagation contribute their own significant reduction in AUFC score. The AUFC reduction due to neuralization alone can be explained by the latter desaturating the original SVM function, stripping it from its local extrema near the training data. The additional AUFC reduction due to propagation can be attributed to the ability of our LRP rules to provide further contextualization (through a large value of $\eta$), which is essential to account for the SVM's overall nonlinear response.

\section{Use Case 1: Explaining Wine Quality}
\label{sec:wine_quality}

In this use case, we revisit the analysis by \cite{CORTEZ2009} on the UCI Wine Quality dataset, which contains 1,599 red and 4,894 white samples of Portuguese Vinho Verde wine. Each sample includes physicochemical measurements (e.g.\ pH, sulfur content) and a quality score (1–10) determined by the median rating from at least three blind tasters. 

In their original analysis, the authors conducted a sensitivity analysis to identify the most influential features in explaining the variance of Gaussian SVM predictions. However, without specialized XAI methods, their approach did not reveal explicit relationships between input features and wine quality, leaving unclear which attributes contribute positively or negatively to perceived taste. Hence, we apply our LRP-SVM method to uncover these directed relationships.

We focused on investigating white wine samples, the larger of the two subsets, and considered their classification into quality ratings above and below 5. We trained one SVM with a low $\gamma$ value of $10^{-4}$ and another with a higher value of $0.2$. The regularization parameter $C$ was set by performing a grid search and optimizing model accuracy. LRP-SVM parameters were set according to our heuristic in Table \ref{tab:heuristic}. As explaining predictions for training data points would incur overfitting biases, we repeatedly split the dataset into training and test sets across five folds. For each split, we trained the SVM on the training data and generated explanations for predictions on the test data. This approach allowed us to obtain explanations for every data point in the dataset.

\begin{figure*}[t!]
    \centering
    \makebox[\textwidth][c]{
    \includegraphics[width=1.1\textwidth]
    {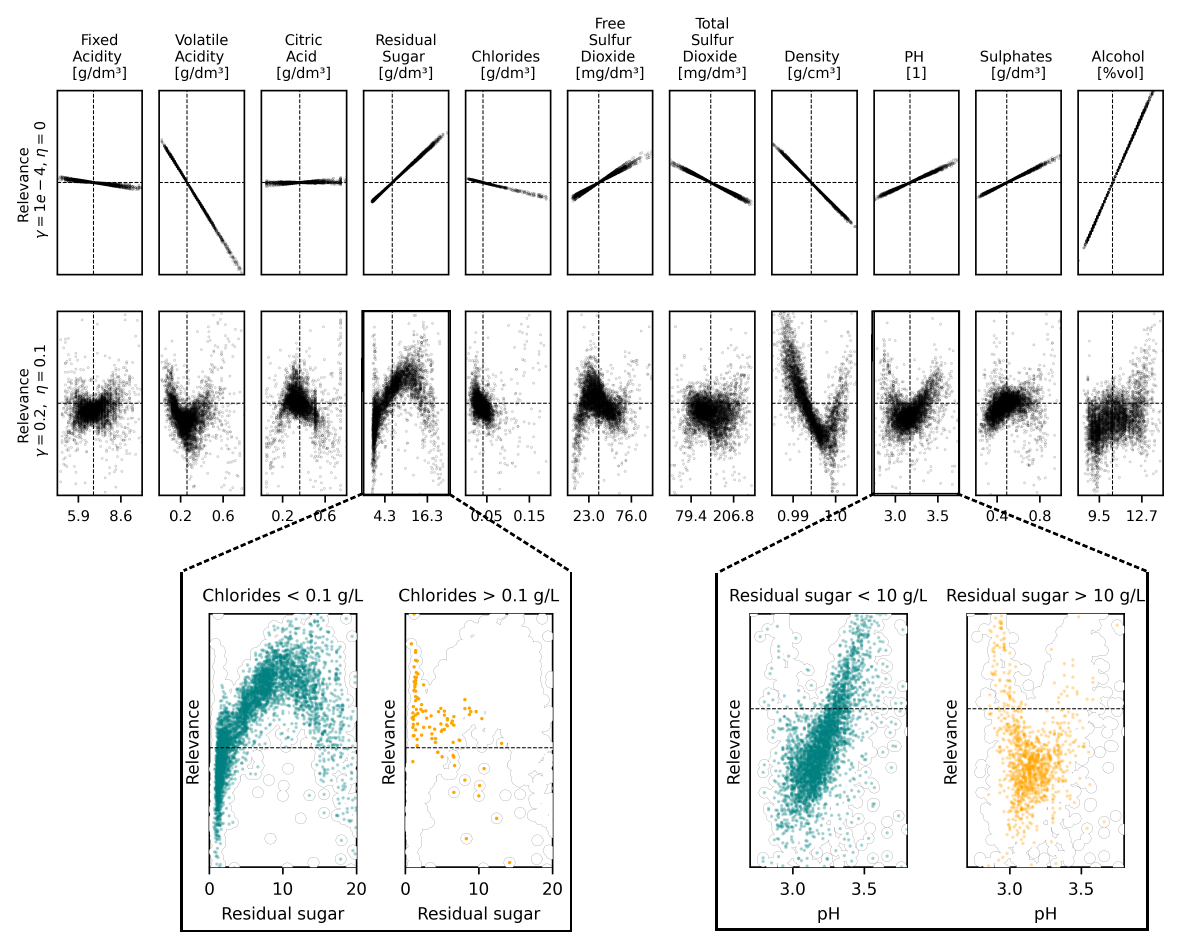}
    }
    \caption{Scatter plots assessing the relevance of chemical properties to high wine quality as inferred by our LRP-SVM method. The first row explores the input-output relation through a low-$\gamma$ SVM while the second row considers a high-$\gamma$ SVM. Vertical dashed lines indicate dataset-wide average feature values and horizontal dashed lines mark zero relevance. At the bottom of the figure, sugar relevance is analyzed based on high/low chloride content, and pH relevance based on high/low sugar content. Our analysis highlights feature interactions, suggesting that chloride content determines the relationship between sugar and taste, while high sugar improves the effect of acidity on taste.}
   \label{fig_wine_relevance_scatter}
\end{figure*}

Figure \ref{fig_wine_relevance_scatter} shows the output of our analysis, specifically the relationship between input values and their contribution to the model output. When the data are modeled by a low-$\gamma$ SVM, our analysis reveals essentially linear connections between features and wine taste. When the data are modeled by a more expressive high-$\gamma$ SVM, our analysis unlocks a richer set of nonlinear associations. For example, when considering the relation between sugar content and taste, our analysis suggests that wines with moderate residual sugar generally taste better, whereas deviations from these levels are less favorable.

A more refined view of sugar's role emerges when we consider chloride levels. In low-chloride wines, sugar shows a negative quadratic relationship to quality -- intermediate levels are most beneficial -- while in high-chloride wines, increasing sugar systematically lowers quality. One possible explanation is that sugar may enhance the perception of saltiness in high-chloride wines \cite{BenAbu2018, Shen2022}, thereby diminishing taste. A similar interaction appears between pH and sugar: while acidity benefits sweet wines, it seems detrimental in wines with low sugar. This effect likely stems from the known effect of sweetness and acidity counterbalancing each other \cite{ZAMORA2006}, creating a more moderate flavor.

\section{Use Case 2: Explaining Dipole Moment Prediction}\label{sec:dpm}

In this second use case, we consider a quantum chemistry application where we explore how the geometric arrangement of a molecule (atomic positions and nuclear charges) affects the distribution of positive and negative charges, as measured by the dipole moment (specifically, its norm). While modeling this type of relation falls within the scope of a regression task rather than classification, the relation can be predicted well using a model structurally similar to the Gaussian SVM in Section \ref{section:neuralization}, namely a function of the type
\begin{equation} \label{eq:krr_pred}
    f(\bx) = \sum_{l=1}^n\alpha_\ell\exp(-\gamma\cdot\|\bx-\bx_\ell\|^2)
\end{equation}
where $\bx$ is a feature representation of the molecule, $f(\bx)$ is a real-valued prediction of the dipole moment, and the $\alpha_\ell$'s can take both positive and negative values and can be optimized using e.g.,\ Kernel Ridge Regression (KRR). In particular, Eq.\ \eqref{eq:krr_pred} can be rewritten in the form of Eq.\ \eqref{eq:svm-boundary} by setting $(\alpha_\ell,y_\ell) \gets (|\alpha_\ell|,\mathrm{sign}(\alpha_\ell))$, which makes it amenable to exactly the same neuralization-propagation procedure described in Sections \ref{section:neuralization} and \ref{section:explanation}. Consequently, level sets of the regression model become robustly attributable to the input features using our LRP-SVM method.

We proceed with our chemistry use case, using as a starting point the QM9 dataset \cite{Wu2018} (130,000 molecules represented by atomic positions, nuclear charges, and target properties such as dipole moment norm) and training KRR models to predict the dipole moment norm. To emphasize the effect of the geometry of the molecule (rather than size) on charge distribution, we constrain our analysis to dipole moment prediction for a specific set of structural isomers, here, C$_7$H$_{10}$O$_2$ and C$_7$H$_{11}$NO for which $6094$ and $5858$ different configurations exist in the dataset. We consider two common representations of molecular geometry: First, a Coulomb matrix (CM) \cite{rupp_2012} representing each pair of atoms by their product of nuclear charges divided by their interatomic distances ($C_{ij} = Z_i Z_j / \|r_i - r_j\|$ for $i \neq j$ and $C_{ii} = Z_i^{2.4}/2$). The rows and columns of the resulting matrix are sorted according to their L1 norm. Second, we consider a Bag of Bonds (BOB) representation \cite{Hansen2015} that partitions interatomic Euclidean distances by atom pair type and records them as sorted lists. Both representations satisfy the desired invariances -- specifically, to rotations, translations, and atom indexing.

\begin{figure*}[t!]
    \centering
    \makebox[\textwidth][c]{
    \includegraphics[width=\textwidth]{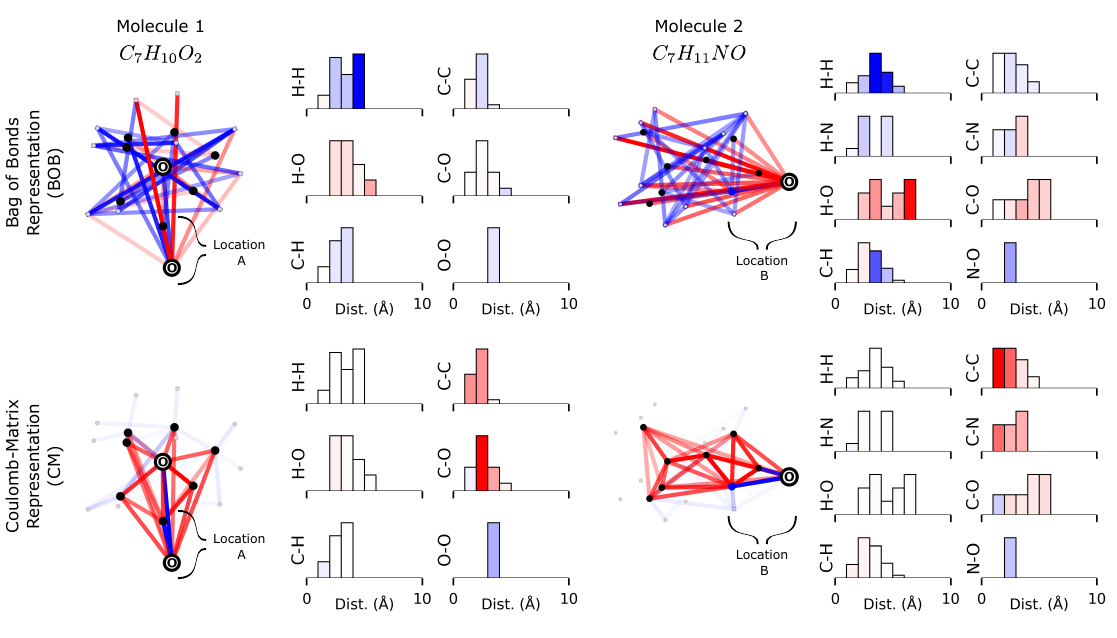}
    }\vspace{-6mm}
    
    \caption{LRP explanations of KRR dipole moment predictions. The first row shows explanations using the Bag of Bonds representation, and the second row uses the Coulomb matrix representation. Each column presents one structural isomer molecule with a high dipole moment norm. On the left of each panel, atom positions are shown as colored circles (hydrogen in grey, carbon in black, oxygen highlighted as white circles marked with ``O''  and nitrogen in blue) with interatomic distance relevance highlighted in red (positive contribution to the dipole moment) and blue (negative contribution). On the right, histograms display interatomic distances grouped by atom pair type, with aggregated relevance overlaid in shades of red and blue. Both representations (CM and BOB) reveal a clear directional pattern driven by peripheral oxygen atoms (Location A in Molecule 1 and Location B in Molecule 2). BOB highlights medium- to long-range interactions across all atom pairs, whereas the CM emphasizes shorter-range distances involving atoms of high nuclear charge.}
    \label{fig:dipole_moment_samples}
\end{figure*}

We use an 80/20 train-test split and train one KRR model for each isomer set using both CM and BOB input representations. In each case, we set $\gamma=3$ and $C = 10$. BOB achieved lower mean absolute errors (0.465 and 0.403 Debye) compared to CM (0.513 and 0.477 Debye). LRP parameters are set according to the heuristic in Table \ref{tab:heuristic}. Figure \ref{fig:dipole_moment_samples} shows explanations of KRR predictions for high-dipole-moment samples from each partition, using both CM and BOB representations. Besides relevance values associated with interatomic distances, Figure \ref{fig:dipole_moment_samples} also displays relevances aggregated to histogram bins of interatomic distances.

The explanations reveal that positive dipole moment relevance aligns with a specific direction in both molecules (vertical in Molecule 1, horizontal in Molecule 2) in both CM and BOB representations.  In particular, both agree on the importance of oxygen at the molecular periphery (Locations A and B in Molecules 1 and 2, respectively). From a basic electrostatic perspective, this is easily attributed to the charge difference between oxygen and the other elements in the molecule, leading to a negative partial charge in favor of oxygen. This effect is particularly visible in the BOB representation, where negative and positive relevances connect clusters of atoms. This reveals how the KRR modeled two distinct charge centers dominating the dipole moment. 
Such an emphasis on two learned charge centers aligns with previous observations in \cite{Schnake2022}. 
From the chemical point of view, moving beyond diatomic molecules introduces ambiguity regarding the direction and magnitude of the dipole moment in larger molecules. 
An intuitive simplification is achieved by clustering atoms until only two groups remain, hence reducing the complicated problem to a 'diatomic' approximation. 

At a more technical level, we can observe significant differences in strategy between KRRs based on different data representations. While the BOB representation relies mainly on long-range interactions, the CM representation is clearly biased in favor of short distances with elements of high nuclear charge (i.e.\ everything but hydrogen). 
This can be explained by the CM being dominated by high energies between nearby atoms, and being multiplied by the atoms' nuclear charges.

Our findings are also clearly identifiable in the histogram representation of these explanations (cf.\ Figure \ref{fig:dipole_moment_samples}), where the CM model is dominated by contributions from small distances (left part of the histograms) with elements of high nuclear charge (C-C, C-O, C-N), while BOB favors large distances (right part of the histograms) between binding partners (e.g.\ C-H, O-H).

In summary, our explanations reveal that the KRR model learned chemically interpretable strategies for dipole moment prediction: identifying as relevant long-range interactions between peripheral locations of the molecule dominated by oxygen atoms. Our analysis also reveals subtle differences in strategy when the representation inhibits those long-range interactions, with CM preserving the overall directionality of relevance but relying instead on short-range interactions. 

\section{Conclusion}

Distance-based classifiers, such as SVMs and KNNs, remain fundamental tools in machine learning due to their ability to efficiently find global optima, their orderly injection of nonlinearity, and their small number of hyperparameters. However, their tendency to exhibit local extrema in input space is a significant hindrance to interpreting the learned prediction strategy.

The present paper addresses this limitation by introducing a reformulation of distance-based classifiers as formally equivalent neural networks with built-in explanatory structures. Specifically, the newly derived neural networks invoke a pooling-detection structure that better conveys the directionality of the decision function and eliminates the problematic local extrema. We furthermore leverage these explanatory structures through novel LRP rules that produce robust explanations under different degrees of model nonlinearity. Our resulting methods, LRP-SVM and LRP-KNN, are computationally efficient, requiring the equivalent of only two evaluations of the original classifier.

Through a comprehensive quantitative evaluation across diverse datasets and models, we demonstrated that our approach consistently delivers robust explanations, outperforming explanation baselines, especially in highly nonlinear models. An ablation study demonstrated that neuralization, by itself, improves model explainability while invoking our proposed LRP rules achieves additional improvement.

Finally, two exploratory use cases demonstrated the practical utility of our approach: explaining an SVM model of wine quality revealed complex nonlinear interactions between chemical wine properties influencing taste. Explanations of a molecular dipole moment regressor highlighted directional features aligning with chemical intuition while revealing biases introduced by the choice of input representation.

\section*{Acknowledgements}
This work was in part supported by the Federal German Ministry for Education and Research (BMBF) under Grants BIFOLD24B, BIFOLD25B, 01IS18037A, 01IS18025A, and 01IS24087C. K.R.M.\ was partly supported by the Institute of Information \& Communications Technology Planning \& Evaluation (IITP) grants funded by the Korea government (MSIT) (No. 2019-0-00079, Artificial Intelligence Graduate School Program, Korea University and No. 2022-0-00984, Development of Artificial Intelligence Technology for Personalized Plug-and-Play Explanation and Verification of Explanation).  Moreover, this work was supported by BASLEARN - TU Berlin/BASF Joint Laboratory, co-financed by TU Berlin and BASF SE.

\bibliographystyle{elsarticle-num}
\bibliography{references.bib}



\section*{Supplementary material}

# Fast and Accurate Explanations of Distance-Based Classifiers by Uncovering Latent Explanatory Structures

(SUPPLEMENTARY NOTES)

Florian Bley, Jacob Kauffmann, Simon León Krug, Klaus-Robert Müller, Grégoire Montavon

This document provides supplementary materials to the main paper. It contains detailed derivations, proofs, and additional technical notes supporting the methods and results presented in the main text.

**Supplementary Note A: Neuralizing Kernel SVMs**

This note derives the neuralized predictive form of Eqs. (2a)–(2c) from the main text, providing a reformulation of the Kernel SVM decision function in terms of soft-pooling operations and linear neurons. We establish an equivalence between the standard kernelized decision rule and its neuralized counterpart, and show that both yield the same classification outcomes.

Recall from the main text the Gaussian kernel:

$$k(\boldsymbol{x}, \boldsymbol{x}') = \exp(-\gamma \cdot \|\boldsymbol{x} - \boldsymbol{x}'\|^2), \tag{1}$$

the smin and smax pooling:

$$\mathrm{smin}_i^\gamma \{a_i\} = -\gamma^{-1} \log\Big( \sum_i \exp(-\gamma \cdot a_i) \Big), \tag{2}$$

$$\mathrm{smax}_i^\gamma \{a_i\} = \gamma^{-1} \log\Big( \sum_i \exp(\gamma \cdot a_i) \Big), \tag{3}$$

and the linear neurons:

$$z_{ij} = (\boldsymbol{x} - \boldsymbol{m}_{ij})^\top \boldsymbol{w}_{ij} + b_{ij} \tag{4}$$

with $\boldsymbol{m}_{ij} = \tfrac{1}{2}(\boldsymbol{u}_i + \boldsymbol{u}_j)$, weights $\boldsymbol{w}_{ij} = 2 \cdot (\boldsymbol{u}_i - \boldsymbol{u}_j)$ and biases $b_{ij} = -\gamma^{-1} \log(\alpha_i/\alpha_j)$.

**Proposition 1** (Equivalence of predictions from Eq. (1) and Eq. (2)). *Let $f(\boldsymbol{x}) = \sum_{i=1}^n y_i \alpha_i k(\boldsymbol{u}_i, \boldsymbol{x}) + \theta$ and $g(\boldsymbol{x}) = \mathrm{smin}_{j \in C^+}^\gamma \{ \mathrm{smax}_{i \in C^-}^\gamma \{z_{ij}\}\}$. For all inputs $\boldsymbol{x} \in \mathcal{X}$ it holds:*

$$\mathrm{sign}(f(\boldsymbol{x})) = \mathrm{sign}(g(\boldsymbol{x})). \tag{5}$$

*In other words, $f(\boldsymbol{x})$ and $g(\boldsymbol{x})$ lead to the same classification for all $\boldsymbol{x}$.*

*Proof.* We show that $f(\boldsymbol{x}) > 0$ if and only if $g(\boldsymbol{x}) > 0$. The same proof steps can be carried out to also show $f(\boldsymbol{x}) = 0$ if and only if $g(\boldsymbol{x}) = 0$.

$$\sum_{i=1}^n y_i \alpha_i k(\boldsymbol{u}_i, \boldsymbol{x}) + \theta > 0 \quad \Leftrightarrow \quad \mathrm{smin}_{j \in C_-}^\gamma \{\mathrm{smax}_{i \in C_+}^\gamma \{z_{ij}\}\} > 0 \tag{6}$$

1

Since $y_i \in \{-1, +1\}$, the SVM decision can be expressed as:

$$\sum_{i \in C^+} \alpha_i k(\boldsymbol{u}_i, \boldsymbol{x}) - \sum_{j \in C^-} \alpha_j k(\boldsymbol{u}_j, \boldsymbol{x}) + \theta > 0 \tag{7}$$

where $C^+$ and $C^-$ collect the indices of positive or negative labels respectively. Depending on the sign of $\theta$ we rearrange terms:

$$\text{For } \theta > 0: \quad \sum_{i \in C^+} \alpha_i k(\boldsymbol{u}_i, \boldsymbol{x}) + \theta > \sum_{j \in C^-} \alpha_j k(\boldsymbol{u}_j, \boldsymbol{x}) \tag{8}$$

$$\text{For } \theta < 0: \quad \sum_{i \in C^+} \alpha_i k(\boldsymbol{u}_i, \boldsymbol{x}) \quad > \sum_{j \in C^-} \alpha_j k(\boldsymbol{u}_j, \boldsymbol{x}) + |\theta|. \tag{9}$$

We now replace $k(\boldsymbol{u}_j, \boldsymbol{x}) = \exp(-\gamma \cdot \|\boldsymbol{u}_j - \boldsymbol{x}\|^2)$. Since both sides are positive, we can take the rescaled negative log on both sides, which flips the inequality. We prove only the case $\theta > 0$ and note that $\theta < 0$ proceeds similarly:

$$-\gamma^{-1} \log \left( \sum_{i \in C^+} \alpha_i \exp(-\gamma \cdot \|\boldsymbol{u}_i - \boldsymbol{x}\|^2) + \theta \right) < -\gamma^{-1} \log \left( \sum_{j \in C^-} \alpha_j \exp(-\gamma \cdot \|\boldsymbol{u}_j - \boldsymbol{x}\|^2) \right). \tag{10}$$

To simplify notation, we define $z_k = \|\boldsymbol{u}_k - \boldsymbol{x}\|^2 - \gamma^{-1} \log \alpha_k$ and $z_0 = -\gamma^{-1} \log \theta$ and include $z_0$ as a constant neuron in the positive set by redefining $C^+ \leftarrow C^+ \cup \{0\}$. This re-parametrization integrates $\theta$ into the pooled representation without changing the decision boundary.

$$-\gamma^{-1} \log \left( \sum_{i \in C^+} \exp(-\gamma \cdot z_i) \right) < -\gamma^{-1} \log \left( \sum_{j \in C^-} \exp(-\gamma \cdot z_j) \right) \tag{11}$$

$$\Leftrightarrow \quad \smin_{i \in C^+}^\gamma \{z_i\} < \smin_{j \in C^-}^\gamma \{z_j\} \tag{12}$$

$$\Leftrightarrow \quad 0 < \smin_{j \in C^-}^\gamma \{z_j\} - \smin_{i \in C^+}^\gamma \{z_i\} \tag{13}$$

$$\Leftrightarrow \quad 0 < \smin_{j \in C^-}^\gamma \{z_j - \smin_{i \in C^+}^\gamma \{z_i\}\} \tag{14}$$

$$\Leftrightarrow \quad 0 < \smin_{j \in C^-}^\gamma \{\smax_{i \in C^+}^\gamma \{z_j - z_i\}\} \tag{15}$$

$$\Leftrightarrow \quad 0 < \smin_{j \in C^-}^\gamma \{\smax_{i \in C^+}^\gamma \{z_{ij}\}\} \tag{16}$$

More details on how we get from Eq. (15) to Eq. (16) can be found in Supplementary Note B. With these transformations, we have shown that

$$f(\boldsymbol{x}) > 0 \Leftrightarrow g(\boldsymbol{x}) > 0. \tag{17}$$

Thus, for all $\boldsymbol{x} \in \mathcal{X}$, the classification outcome based on $\text{sign}(f(\boldsymbol{x}))$ matches the classification outcome based on $\text{sign}(g(\boldsymbol{x}))$. □

These transformations demonstrate that the decision boundary of the neuralized formulation is mathematically equivalent to that of the original Kernel SVM. This result confirms that the classification behavior remains unchanged under the proposed neuralization.

## Supplementary Note B: Neuralizing KNN Classifier

This supplementary note contains derivations and proofs for the neuralized KNN classifier, providing a rigorous justification of the equivalence results discussed in the main text. Recall from the main text the ranked min and max operator:

$$\text{rmin}_i^q\{a_i\} = \min_i \{a_i : \sum_j I(a_j \leq a_i) \geq q\}, \tag{18}$$

$$\text{rmax}_i^q\{a_i\} = \max_i \{a_i : \sum_j I(a_j \geq a_i) \geq q\}, \tag{19}$$

and the linear neurons:

$$z_{ij} = (\boldsymbol{x} - \boldsymbol{m}_{ij})^\top \boldsymbol{w}_{ij} \tag{20}$$

with $\boldsymbol{m}_{ij} = \frac{1}{2}(\boldsymbol{u}_i + \boldsymbol{u}_j)$ and weights $\boldsymbol{w}_{ij} = 2(\boldsymbol{u}_i - \boldsymbol{u}_j)$.

**Proposition 2** (Equivalence of predictions from Eq. (3) and Eq. (4)). *Let $f(\boldsymbol{x}) = \sum_{i=1}^n y_i I(\boldsymbol{u}_i \in N_K(\boldsymbol{x}))$ and $g(\boldsymbol{x}) = \text{rmin}_{j \in C^-}^q \{\text{rmax}_{i \in C^+}^q\{z_{ij}\}\}$. For all inputs $\boldsymbol{x} \in \mathcal{X}$ it holds:*

$$\text{sign}(f(\boldsymbol{x})) = \text{sign}(g(\boldsymbol{x})). \tag{21}$$

*In other words, $f(\boldsymbol{x})$ and $g(\boldsymbol{x})$ lead to the same classification for all $\boldsymbol{x}$.*

*Proof.* Let $d_K = \|\boldsymbol{x} - \boldsymbol{u}_{(K)}\|$ be the distance to the $K$-nearest neighbor of $\boldsymbol{x}$. Assume $\boldsymbol{x}$ is assigned to the positive class. We will show the following equivalence:

$$\sum_i y_i I(\|\boldsymbol{x} - \boldsymbol{u}_i\| \leq d_K) > 0 \quad \Leftrightarrow \quad \text{rmin}_{j \in C^-}^q\{\text{rmax}_{i \in C^+}^q\{z_{ij}\}\} > 0. \tag{22}$$

Since $y_i \in \{-1, +1\}$, the KNN decision can be expressed as:

$$\sum_{i \in C^+} I(\|\boldsymbol{x} - \boldsymbol{u}_i\| \leq d_K) > \sum_{j \in C^-} I(\|\boldsymbol{x} - \boldsymbol{u}_j\| \leq d_K) \tag{23}$$

Let $q = (k+1)/2$, and observe that the lhs is at least $q$ while the rhs is at most $q - 1$. Applying the definition of rmin yields the following equivalences:

$$\sum_{i \in C^+} I(\|\boldsymbol{x} - \boldsymbol{u}_i\| \leq d_K) \geq q \quad \Leftrightarrow \quad \text{rmin}_{i \in C^+}^q\{\|\boldsymbol{x} - \boldsymbol{u}_i\|\} \leq d_K \tag{24}$$

$$\sum_{j \in C^-} I(\|\boldsymbol{x} - \boldsymbol{u}_j\| \leq d_K) < q \quad \Leftrightarrow \quad \text{rmin}_{j \in C^-}^q\{\|\boldsymbol{x} - \boldsymbol{u}_j\|\} > d_K \tag{25}$$

Combining the two inequalities of $\text{rmin}_i$ and $\text{rmin}_j$ yields:

$$\text{rmin}_{i \in C^+}^q\{\|\boldsymbol{x} - \boldsymbol{u}_i\|\} < \text{rmin}_{j \in C^-}^q\{\|\boldsymbol{x} - \boldsymbol{u}_j\|\} \tag{26}$$

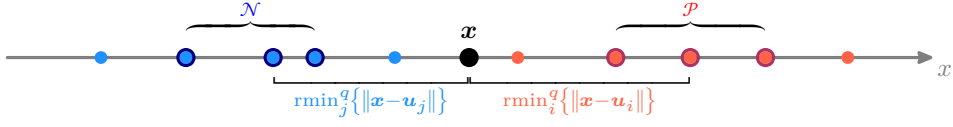

Figure 1: A test sample $x$ is classified to belong to the blue class because its $q$th neighbor ($q = 3$) is closer than the $q$th neighbor from the red class. The two sets $\mathcal{N}$ and $\mathcal{P}$ are constructed by adding the $\kappa = 1$ closer and further neighbors to the sets, as well as the $q$th neighbor itself.

We now simplify as follows:

$$\operatorname*{rmin}_{i \in C^+}^q \{\|x - u_i\|^2\} < \operatorname*{rmin}_{j \in C^-}^q \{\|x - u_j\|^2\} \tag{27}$$

$$\Leftrightarrow \quad 0 < \operatorname*{rmin}_{j \in C^-}^q \{\|x - u_i\|^2\} - \operatorname*{rmin}_{i \in C^+}^q \{\|x - u_i\|^2\} \tag{28}$$

$$\Leftrightarrow \quad 0 < \operatorname*{rmin}_{j \in C^-}^q \{\|x - u_j\|^2 - \operatorname*{rmin}_{i \in C^+}^q \{\|x - u_i\|^2\}\} \tag{29}$$

$$\Leftrightarrow \quad 0 < \operatorname*{rmin}_{j \in C^-}^q \{\operatorname*{rmax}_{i \in C^+}^q \{\|x - u_j\|^2 - \|x - u_i\|^2\}\} \tag{30}$$

$$\Leftrightarrow \quad 0 < \operatorname*{rmin}_{j \in C^-}^q \{\operatorname*{rmax}_{i \in C^+}^q \{-2x^\top u_j + \|u_j\|^2 + 2x^\top u_i - \|u_i\|^2\}\} \tag{31}$$

$$\Leftrightarrow \quad 0 < \operatorname*{rmin}_{j \in C^-}^q \{\operatorname*{rmax}_{i \in C^+}^q \{(x - m_{ij})^\top w_{ij}\}\} \tag{32}$$

$$\Leftrightarrow \quad 0 < \operatorname*{rmin}_{j \in C^-}^q \{\operatorname*{rmax}_{i \in C^+}^q \{z_{ij}\}\} \tag{33}$$

With these transformations, we have shown that

$$f(x) > 0 \Leftrightarrow g(x) > 0. \tag{34}$$

Thus, for all $x \in \mathcal{X}$, the classification outcome based on $\operatorname{sign}(f(x))$ matches the classification outcome based on $\operatorname{sign}(g(x))$. □

This completes the proof of equivalence, establishing that the neuralized formulation retains the same decision boundary as the original KNN classifier.

**Supplementary Note C: Factorization of relevance in nested pooling layers**

This supplementary note shows that for both, KNN and Kernel SVMs that intermediate relevances can be factorized as:

$$R_{ij} = p_i \cdot p_j \cdot g(x) \tag{35}$$

where it can be shown that $p_i$ is identical for all $j$. In the case of a neuralized Kernel SVM we observe:

$$p_i = \frac{\exp(\beta z_{ij})}{\sum_{i'} \exp(\beta z_{i'j})} = \frac{\exp(\beta(z_j - z_i))}{\sum_{i'} \exp(\beta(z_j - z_{i'}))} = \frac{\exp(-\beta z_i)}{\sum_{i'} \exp(-\beta z_{i'})}. \tag{36}$$

It follows that $p_i$ is equivalent for all $j$. In the case of the neuralized KNN we define the following "band of indices":

$$\mathcal{N} = \left\{ j \in C_- \mid q - \kappa \leq \text{rank}_{j'}(h_j) \leq q + \kappa \right\} \tag{37}$$

$$\mathcal{P}_j = \left\{ i \in C_+ \mid q - \kappa \leq \text{rank}_{i'}(z_{ij}) \leq q + \kappa \right\} \tag{38}$$

$$= \left\{ i \in C_+ \mid q - \kappa \leq \text{rank}_{i'}(\|\boldsymbol{x} - \boldsymbol{u}_i\|^2) \leq q + \kappa \right\} \tag{39}$$

$$= \mathcal{P} \tag{40}$$

Since $p_i$ is always constructed from the same set $\mathcal{P}$, it follows again, that $p_i$ is equivalent for all $j$ (and since the redistribution is uniform, also equivalent for all $i \in \mathcal{P}$). Figure 1 depicts visually how the sets $\mathcal{N}$ and $\mathcal{P}$ are constructed from two independent distance rankings.

**Supplementary Note D: Dealing with the Intercept**

This section details the integration of the SVM intercept $\theta$ into the neuralized model and its impact on the LRP framework. By introducing a constant neuron $z_0$, we ensure that the intercept term is properly accounted for within the neural architecture. We further establish how this formulation maintains the affine structure required for consistent application of the relevance model.

The most direct way to incorporate the SVM intercept is to place a "virtual" support vector at the data point $\boldsymbol{x}$ itself:

$$f(\boldsymbol{x}) = \sum_{\ell=1}^{n} y_\ell \cdot \alpha_\ell \exp(-\gamma \cdot \|\boldsymbol{x} - \boldsymbol{u}_\ell\|^2) + \theta \tag{41}$$

$$= \sum_{\ell=1}^{n} y_\ell \cdot \alpha_\ell \exp(-\gamma \cdot \|\boldsymbol{x} - \boldsymbol{u}_\ell\|^2) + \underbrace{\text{sign}(\theta)}_{=y_0} \cdot \underbrace{|\theta|}_{=\alpha_0} \exp(-\gamma \cdot \underbrace{\|\boldsymbol{x} - \boldsymbol{x}\|^2}_{=\boldsymbol{u}_0}) \tag{42}$$

$$= \sum_{\ell=0}^{n} y_\ell \cdot \alpha_\ell \exp(-\gamma \cdot \|\boldsymbol{x} - \boldsymbol{u}_\ell\|^2) \tag{43}$$

where the sum now starts at 0 instead of 1, and the intercept $\theta$ is consumed by the term $y_0 \cdot \alpha_0$. By treating the "virtual" support vector $\boldsymbol{u}_0$ as constant (although different for every data point $\boldsymbol{x}$), the prediction can be neuralized and explained exactly as described in the main text without any further modifications. Below, we provide additional information for how this simple step seamlessly integrates into our framework. In particular, we show how the propagation rules redistribute the intercept onto intermediate neurons and input variables.

Recall that the SVM intercept $\theta$ is integrated into the neuralized model by introducing a constant neuron:

$$z_0 = -\gamma^{-1} \log |\theta|. \tag{44}$$

This constant neuron $z_0$ is added to the pool corresponding to the class favored by the intercept – that is, to the $C^+$ pool when $\theta > 0$, or the $C^-$ pool when $\theta < 0$. This inclusion ensures that the intercept term is properly accounted for in the neural network architecture.

Eqs. (15) and (16) establish the relationship between the squared Euclidean distances with the linear layers of the neuralized SVM:

$$z_j - z_i = [\|x - u_j\|^2 - \log \alpha_j] - [\|x - u_i\|^2 - \log \alpha_i] \tag{45}$$

$$= (x - m_{ij})^\top w_{ij} + b_{ij} \tag{46}$$

$$= z_{ij} \tag{47}$$

In our proposed relevance model,

$$\widehat{R}_{ij} = [p_i p_j]_{\text{cst}} \cdot z_{ij}, \tag{48}$$

we treat $\widehat{R}_{ij}$ as an affine transformation of $z_{ij}$, which consequently becomes an affine transformation of $x$ when $z_{ij}$ is affine.

The inclusion of the constant neuron $z_0$ requires careful handling to maintain the affine structure of the relevance model. Specifically, for neurons involving $z_0$, the quadratic terms in $x$ do not cancel out as they do in the difference $z_j - z_i$. To maintain the affine structure of the relevance model for these neurons, we introduce the "virtual support vector" $u_0$ located at the data point $x$:

$$z_0 = \|x - u_0\|^2 - \gamma^{-1} \log |\theta| \tag{49}$$

where $u_0 = x$. Although this effectively adds zero to the expression, treating $u_0$ as constant restores the affine dependence of the relevance model on $x$ for neurons involving $z_0$.

By incorporating $u_0 = x$, the expressions for $z_{0j}$ and $z_{i0}$ become affine functions of $x$, allowing us to apply the relevance model uniformly across all neurons. For $\theta > 0$, we obtain:

$$z_{0j} = [\|x - u_j\|^2 - \gamma^{-1} \log \alpha_j] - [\|x - u_0\|^2 - \gamma^{-1} \log \theta] \tag{50}$$

$$= (x - m_{0j})^\top w_{0j} + b_{0j} \tag{51}$$

with $m_{0j} = \frac{1}{2}(u_0 + u_j)$, $w_{0j} = 2 \cdot (u_0 - u_j)$ and $b_{0j} = \gamma^{-1} \log \frac{\theta}{\alpha_j}$. Similarly, for $\theta < 0$:

$$z_{i0} = [\|x - u_0\|^2 - \gamma^{-1} \log |\theta|] - [\|x - u_i\|^2 - \gamma^{-1} \log \alpha_i] \tag{52}$$

$$= (x - m_{i0})^\top w_{i0} + b_{i0}. \tag{53}$$

By expressing $z_{0j}$ and $z_{i0}$ in this affine form, we ensure that the relevance model remains consistent across all neurons, including those involving the intercept term, allowing us to apply the $\eta$-rule without modification.

**Supplementary Note E: Deriving efficient LRP explanations**

This supplementary note shows how $\mathcal{E}_\eta(x)$ from Eq. (7) in the main paper can efficiently be computed by interpolating between the two special cases $\mathcal{E}_{\eta=0}$ and $\mathcal{E}_{\eta=1}$. First, we show that $\mathcal{E}_{\eta=0}$ can be computed efficiently:

$$\mathcal{E}_{\eta=0}(x) = \sum_{i \in C_+} \sum_{j \in C_-} x \odot w_{ij} \cdot p_i p_j \tag{54}$$

$$= \sum_{i \in C_+} \sum_{j \in C_-} x \odot 2(u_i - u_j) \cdot p_i p_j \tag{55}$$

$$= x \odot \Big( \sum_{i \in C_+} 2 u_i p_i - \sum_{j \in C_-} 2 u_j p_j \Big) \tag{56}$$

$$= x \odot \nabla g(x), \tag{57}$$

i.e. $\eta = 0$ can be computed efficiently as Gradient × Input on the neuralized SVM. Next, we show that $\mathcal{E}_{\eta=1}$ can be computed efficiently:

$$\mathcal{E}_{\eta=1}(x) = \sum_{i \in C_+} \sum_{j \in C_-} (x - m_{ij}) \odot w_{ij} \cdot p_i p_j \quad (58)$$

$$= \sum_{i \in C_+} \sum_{j \in C_-} ((x - u_j)^2 - (x - u_i)^2) \cdot p_i p_j \quad (59)$$

$$= \sum_{j \in C_-} p_j \cdot (x - u_j)^2 - \sum_{i \in C_+} p_i \cdot (x - u_i)^2, \quad (60)$$

which is efficient since the nested sum factorizes into two independent summations. And finally, we show that interpolating between $\mathcal{E}_{\eta=0}$ and $\mathcal{E}_{\eta=1}$ is equivalent to computing $\mathcal{E}_\eta$:

$$\mathcal{E}_\eta(x) = \sum_{i \in C_+} \sum_{j \in C_-} (x - \eta \cdot m_{ij}) \odot w_{ij} \cdot p_i p_j \quad (61)$$

$$= \sum_{i \in C_+} \sum_{j \in C_-} \left((1 - \eta) \cdot x + \eta \cdot (x - m_{ij})\right) \odot w_{ij} \cdot p_i p_j \quad (62)$$

$$= (1 - \eta) \cdot \sum_{i \in C_+} \sum_{j \in C_-} x \odot w_{ij} \cdot p_i p_j + \eta \cdot \sum_{i \in C_+} \sum_{j \in C_-} (x - m_{ij}) \odot w_{ij} \cdot p_i p_j \quad (63)$$

$$= (1 - \eta) \cdot \mathcal{E}_{\eta=0}(x) + \eta \cdot \mathcal{E}_{\eta=1}(x) \quad (64)$$

Thus, we have shown that $\mathcal{E}_\eta$ can be computed in $O(|C_-| + |C_+|)$ instead of the naive direct implementation in $O(|C_-| \times |C_+|)$.

## Supplementary Note F: Datasets

The following table provides details about the datasets referenced in the main text.

Table 1: Overview of datasets used in the experiments

| Name | # Samples | # Features | URL |
| --- | --- | --- | --- |
| BreastMNIST | 780 | 784 | `https://doi.org/10.5281/zenodo.10519652` |
| Breast Cancer | 286 | 9 | `https://doi.org/10.24432/C51P4M` |
| Car Evaluation | 1728 | 6 | `https://doi.org/10.24432/C5JP48` |
| Catalyst | 300 | 49 | `https://doi.org/10.1021/acscatal.0c04629` |
| Concrete | 1030 | 8 | `https://doi.org/10.24432/C5PK67` |
| Diabetes Risk | 520 | 16 | `https://doi.org/10.24432/C5VG8H` |
| PneumoniaMNIST | 5856 | 784 | `https://doi.org/10.5281/zenodo.10519652` |
| Raisin | 900 | 7 | `https://doi.org/10.24432/C5660T` |
| Real Estate Valuation | 414 | 6 | `https://doi.org/10.24432/C5J30W` |
| Wine Quality | 4898 | 11 | `https://doi.org/10.24432/C56S3T` |

## Supplementary Note G: Empirical Evaluation of LRP Hyperparameters

LRP for distance-based classifiers relies on the hyperparameters $\eta$ and $\beta$ for SVMs and $\kappa$ for KNN. To systematically determine these hyperparameters, this section aims to derive empirical heuristics based on the SVM scale parameter $\gamma$ and the number of neighbors $k$, which govern the model's non-linearity.

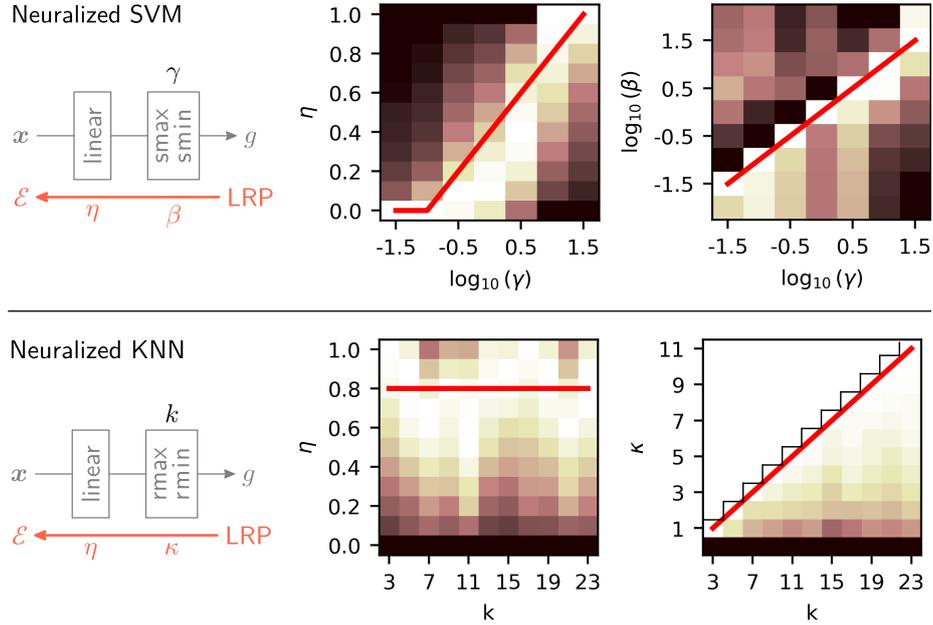

Figure 2: LRP hyperparameter analysis for SVM and KNN. Left panels: schematic of relevance propagation through the network, with LRP parameters assigned to specific layers ($\eta, \beta$ for SVM; $\eta, \kappa$ for KNN). Middle and right panels: color-coded heatmaps of pixel-flipping performance (AUFC values) across combinations of model parameters ($\gamma$ for SVM, $k$ for KNN) and LRP parameters. The middle panels vary the detection unit parameter ($\eta$), while the right panels vary pooling layer parameters ($\beta, \kappa$). Each tile represents the average AUFC for one parameter combination; lighter colors indicate better explanation faithfulness (lower AUFC) with the best AUFC per model parameter (x-axis) coded as white and the worst AUFC coded as black. Red lines represent our heuristics for setting LRP parameters.

Using sklearn's `make_classification` function, we generated 130 binary classification datasets, each with 2,000 samples and 20 features. We varied dataset parameters such as class imbalance, the number of clusters per class, and the proportion of informative features to create diverse datasets.

For both KNN and SVM models, we identified the optimal model hyperparameters via grid search—specifically, the number of neighbors $k$ for KNN and the $\gamma$ parameter for SVM. Consequently, each dataset was associated with a unique optimal $k$ or $\gamma$ parameter. For every value of $k$ or $\gamma$ in our parameter grid, we selected five datasets where that parameter value was optimal.

On these selected datasets, we computed LRP explanations for 100 test points. In doing so, we varied one of the LRP hyperparameters over a defined range—namely, $\eta, \beta$ for SVM, and $\eta, \kappa$ for KNN. We calculated the AUFC for each explanation and then averaged these AUFC values over the 5 datasets corresponding to each optimal model hyperparameter. We depict the resulting AUFC values over the grids of the model ($\gamma, \kappa$) and LRP ($\eta, \beta, \kappa$) hyperparameters in Figure 2.

Our analysis reveals that nearly linear models with low $\gamma$ do not benefit from detection layer contextualization, making $\eta = 0$ the optimal choice. As the RBF kernel's non-linearity increases, higher $\eta$ values become more effective. Additionally, we find no advantage in using a different degree of non-linearity when propagating relevance through the pooling layer beyond what is already dictated by the SVM's $\gamma$ parameter.

For KNN, we observe no strong empirical relationship between $\eta$ and $k$. Instead, a relatively high level of detection layer contextualization performs well across all tested $k$ values. Furthermore, setting $\kappa$ to its maximum admissible value consistently yields optimal explanation accuracy.



\end{document}